\documentclass[lettersize,journal]{IEEEtran}
\usepackage{amsmath,amsfonts}
\usepackage{algorithmic}
\usepackage{algorithm}
\usepackage{array}
\usepackage[caption=false,font=normalsize,labelfont=sf,textfont=sf]{subfig}
\usepackage{textcomp}
\usepackage{stfloats}
\usepackage{url}
\usepackage{verbatim}
\usepackage{graphicx}
\usepackage{cite}
\usepackage{booktabs} 
\usepackage{balance}
\usepackage[breaklinks=true,colorlinks, citecolor=blue]{hyperref}
\begin{document}

\title{DescribeEarth: Describe Anything for Remote Sensing Images}

\author{Kaiyu Li*, Zixuan Jiang*, Xiangyong Cao$^{\dag}$, Jiayu Wang, Yuchen Xiao, Deyu Meng, Zhi Wang
\thanks{This work is partially supported by the National Key R\&D Program of China (2021ZD0112902), and China NSFC projects under contract 62272375, 12226004. \textit{(Corresponding author: Xiangyong Cao)}}
\thanks{Kaiyu Li and Zhi Wang are with School of Software Engineering, Xi’an Jiaotong University, Xi’an 710049, China (email: likyoo.ai@gmail.com, zhiwang@xjtu.edu.cn)}
\thanks{Zixuan Jiang is with College of Artificial Intelligence, Xi’an Jiaotong University, Xi’an 710049, China (email: andrewjiang@stu.xjtu.edu.cn)}
\thanks{Xiangyong Cao, Jiayu Wang, and Yuchen Xiao are with School of Computer Science and Technology and Ministry of Education Key Lab For Intelligent Networks and Network Security, Xi’an Jiaotong University, Xi’an 710049, China (email: caoxiangyong@xjtu.edu.cn, 2234112262@stu.xjtu.edu.cn, 3283879@qq.com)}
\thanks{Deyu Meng is with School of Mathematics and Statistics and Ministry of Education Key Lab of Intelligent Networks and Network Security, Xi’an Jiaotong University, Xi’an, Shaanxi, China, and Pazhou Laboratory (Huangpu), Guangzhou, Guangdong, China. (email: dymeng@mail.xjtu.edu.cn).}
}


\markboth{Journal of \LaTeX\ Class Files,~Vol.~14, No.~8, August~2021}%
{Shell \MakeLowercase{\textit{et al.}}: A Sample Article Using IEEEtran.cls for IEEE Journals}


\maketitle

\begin{abstract}
Automated textual description of remote sensing images is crucial for unlocking their full potential in diverse applications, from environmental monitoring to urban planning and disaster management.
However, existing studies in remote sensing image captioning primarily focus on the image level, lacking object-level fine-grained interpretation, which prevents the full utilization and transformation of the rich semantic and structural information contained in remote sensing images. To address this limitation, we propose Geo-DLC, a novel task of object-level fine-grained image captioning for remote sensing. To support this task, we construct DE-Dataset, a large-scale dataset contains 25 categories and 261,806 annotated instances with detailed descriptions of object attributes, relationships, and contexts. Furthermore, we introduce DE-Benchmark, a LLM-assisted question-answering based evaluation suite designed to systematically measure model capabilities on the Geo-DLC task. We also present DescribeEarth, a Multi-modal Large Language Model (MLLM) architecture explicitly designed for Geo-DLC, which integrates a scale-adaptive focal strategy and a domain-guided fusion module leveraging remote sensing vision-language model features to encode high-resolution details and remote sensing category priors while maintaining global context. Our DescribeEarth model consistently outperforms state-of-the-art general MLLMs on DE-Benchmark, demonstrating superior factual accuracy, descriptive richness, and grammatical soundness, particularly in capturing intrinsic object features and surrounding environmental attributes across simple, complex, and even out-of-distribution remote sensing scenarios. All data, code and weights are released at \url{https://github.com/earth-insights/DescribeEarth}.
\end{abstract}


\section{Introduction}
~\label{sec:intro}

\IEEEPARstart{R}{emote} sensing images, continuously acquired from satellites, aerial platforms, and drones, provide an indispensable tool for monitoring and understanding the Earth. Its applications span environmental science~\cite{chen2022remote, ma2024transfer}, urban planning~\cite{wang2022unetformer, yu2023urban}, disaster management~\cite{casagli2023landslide}, agriculture~\cite{benami2021uniting}, and defense~\cite{harrison2022commercial}. Automatically generating precise and detailed textual descriptions for specific features or phenomena within these images is crucial for unlocking their full potential, enabling more efficient analysis, rapid decision-making, and broader accessibility for both domain experts and the public~\cite{lu2017exploring, cheng2022nwpu, huang2024understanding}.

\begin{figure}
    \centering
    \includegraphics[width=1\linewidth]{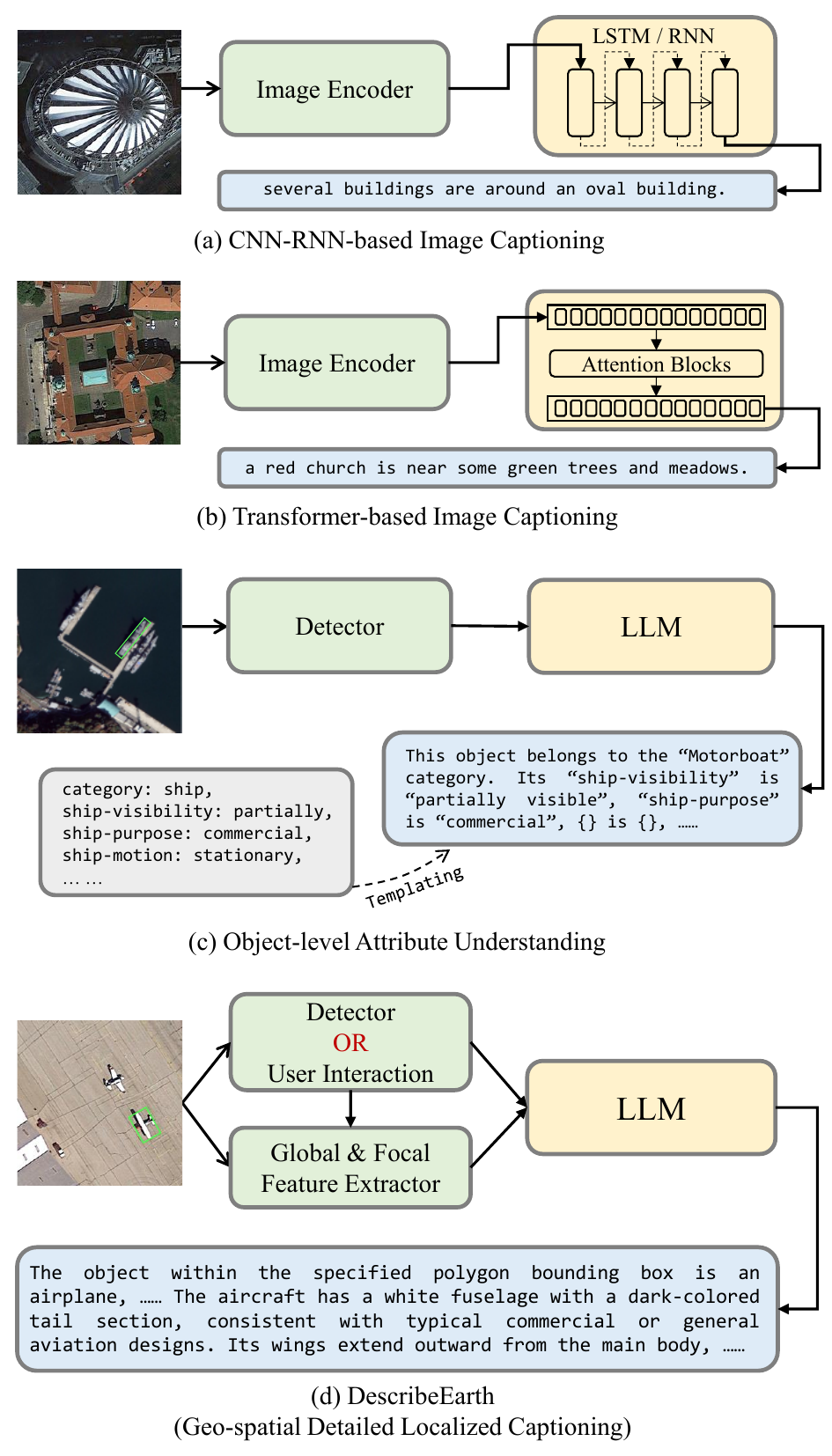}
    \caption{Some methods proposed to resolve remote sensing image description related tasks. (a) and (b) are typical image captioning frameworks~\cite{lu2017exploring, zhang2021shiprsimagenet}. (c) is a recent object description model~\cite{jiang2025eaglevision}, but it can only generate template-based descriptions across a limited number of categories and does not support interactive descriptions. (d) is our DescribeEarth model, which can generate detailed, open-ended and localized descriptions based on off-the-shelf detectors or user interactions.}
    \label{fig:tasks}
\end{figure}

\begin{figure*}
    \centering
    \includegraphics[width=\textwidth]{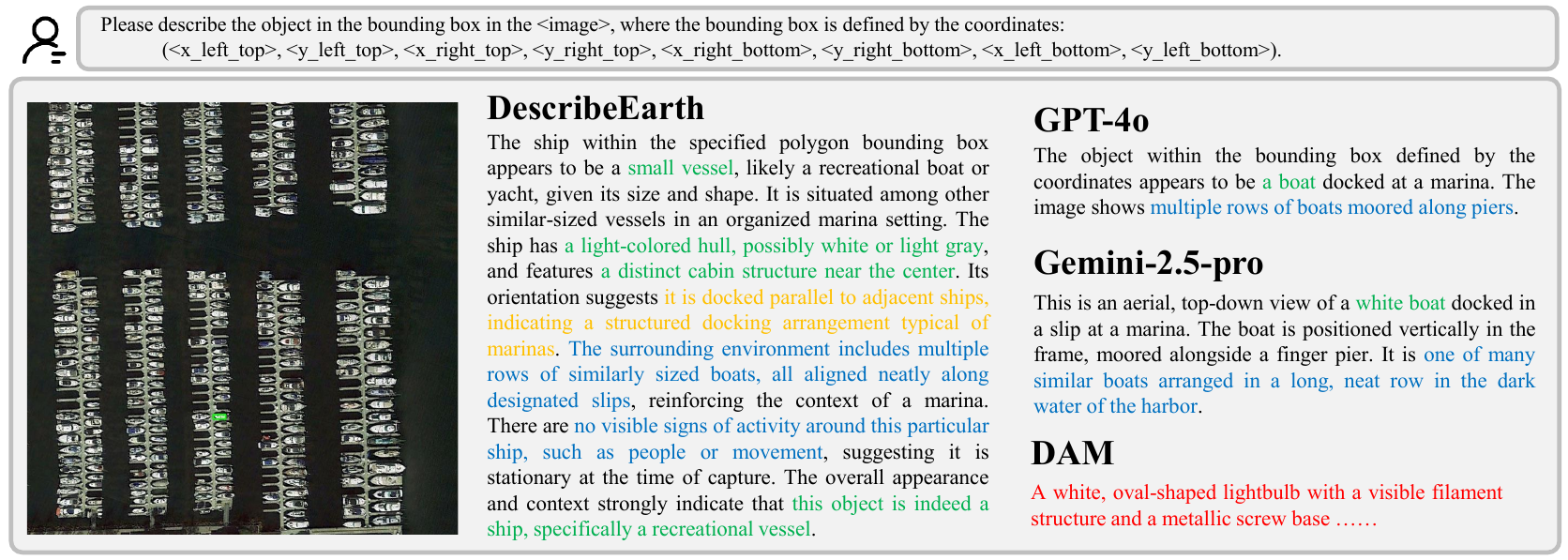}
    \caption{Comparison of DescribeEarth, DAM~\cite{xiao2025describe}, and other general MLLMs~\cite{hurst2024gpt, comanici2025gemini} in describing remote sensing images. \textcolor[RGB]{0,176,80}{Green} rendering represents intrinsic features of the target, \textcolor[RGB]{255,192,0}{yellow} denotes spatial features, \textcolor[RGB]{0,112,192}{blue} indicates contextual features, and \textcolor[RGB]{255,0,0}{red} signifies incorrect descriptions. DescribeEarth delivers the most detailed and accurate object descriptions.}
    \label{fig:motivation_sample}
\end{figure*}

Historically, the task of image description, often referred to as image captioning, primarily focused on generating a single, holistic sentence or paragraph for an entire image~\cite{ghandi2023deep, albadarneh2025attention}. Many methods have been proposed to solve this challenge, and some typical methods are presented in Fig.~\ref{fig:tasks}. Early approaches relied on architectures like CNN-RNN encoders~\cite{zhou2019re, huang2020image, ye2022joint} or attention-based mechanisms~\cite{liu2022remote, chang2023changes, zhang2025intra}, often trained on small datasets and producing brief, coarse-grained sentences. With the advent of Multi-modal Large Language Models (MLLMs), these capabilities have significantly advanced, allowing for more coherent and contextually rich descriptions for general images~\cite{liu2023visual, liu2024improved, bai2025qwen2, abdin2024phi, team2023gemini}. In the remote sensing domain, early attempts at image captioning also focused on scene-level descriptions, often adapting general vision-language models (VLMs) to classify and provide coarse summaries of entire remote sensing images~\cite{kuckreja2024geochat, muhtar2024lhrs, irvin2024teochat, pang2025vhm}. More recently, some efforts have moved towards generating region-specific descriptions. For instance, EagleVision~\cite{jiang2025eaglevision} proposed an object-level attribute MLLM specifically for remote sensing, aiming to provide attribute-rich descriptions for detected objects. While EagleVision represents a step towards localized understanding in remote sensing, it often produces descriptions that are constrained by predefined attributes or templates, limiting the richness and open-ended detail necessary for comprehensive analysis. Moreover, the few training categories and non-interactive operations further limit its practicality.

The recent introduction of the Describe Anything Model (DAM)~\cite{lian2025describe} marks a significant leap in Detailed Localized Captioning (DLC) for natural images and videos. DAM, through its focal prompt and localized vision backbone, adeptly balances local detail with global context, generating meticulously detailed descriptions for user-specified regions. Some recent work follows it and has also made significant contributions to this task~\cite{xiao2025describe, vu2025describe, lin2025perceive}. While DAM and some MLLMs excel in the natural image domain, its direct application to remote sensing images reveals substantial performance gaps. As shown in Fig.~\ref{fig:motivation_sample}, we present the description results of DAM and several general MLLMs on a remote sensing image. GPT-4o~\cite{hurst2024gpt} and Gemini-2.5-pro~\cite{comanici2025gemini} can only provide brief descriptions, lacking details of object attributes, spatial relationships, and contextual information. DAM has difficulty defining categories and providing precise descriptions for some remote sensing instances. This is primarily due to the differences between natural and remote sensing visual data, including unique perspectives (\textit{e.g.}, nadir views), vast scale variations of objects, and the distinct semantic contexts relevant to geographical analysis. Consequently, general models like DAM, trained on natural image characteristics, struggle to accurately identify and provide detailed descriptions of objects and phenomena observed from remote sensing perspectives.

To bridge this critical gap and enable DLC for remote sensing images, a task we formally define as Geo-DLC, we need to address some challenges. Based on our analysis and building upon the inherent difficulties of localized captioning~\cite{lian2025describe}, we identify three obstacles for Geo-DLC:

\begin{itemize}

\item The development of Geo-DLC models is constrained by the lack of dedicated datasets. Existing remote sensing datasets typically offer labels for classification~\cite{xia2017aid}, bounding boxes for detection~\cite{xia2018dota}, masks for segmentation~\cite{wang2021loveda}, or coarse descriptions for the entire image captioning~\cite{qu2016deep, lu2017exploring, cheng2022nwpu, liu2022remote}, but rarely provide the instance-level textual descriptions required for Geo-DLC. Manually creating such a dataset is impractical, requiring substantial resources and specialized geospatial expertise to accurately describe complex details and features.

\item General VLMs, trained on natural images, possess limited recognition capabilities for remote sensing targets~\cite{liu2024remoteclip, zhang2024rs5m, wang2024skyscript, klemmer2025satclip}. Objects in remote sensing images appear from unique perspectives, exhibit vast scale variations, and contain a significant number of small targets~\cite{rolf2024position}. This makes them different from objects in natural images. Consequently, general models struggle to accurately identify and interpret even common objects when viewed from these specialized remote sensing perspectives, let alone capture their intricate details.

\item Beyond training data, there is a lack of an evaluation benchmark specifically designed for Geo-DLC. As stated in~\cite{lian2025describe}, traditional language metrics~\cite{papineni2002bleu, lin2004rouge, banerjee2005meteor} and current LLM-based judgments~\cite{gu2024survey} unfairly penalize models for generating correct details not present in an often incomplete reference caption.

\end{itemize}

To address these limitations and unlock the potential of detailed localized understanding in remote sensing, this paper introduces a comprehensive practice for Geo-DLC. Our contributions are summarized as follow:

\begin{itemize}

\item We introduce DE-Dataset, the first dataset for Geo-DLC. This dataset contains oriented bounding boxes (OBBs) for a diverse range of geographical objects, paired with instance-level detailed textual descriptions. DE-Dataset is constructed through a well designed data pipeline that leverages MLLMs and existing remote sensing object detection datasets, assisted by human verification. This enables efficient scaling of annotation to vast amounts of remote sensing images.

\item We present DescribeEarth, a MLLM architecture explicitly designed for Geo-DLC task. To address the limited recognition capability of general models for remote sensing targets, DescribeEarth utilizes RemoteCLIP's features as a guiding prior~\cite{liu2024remoteclip} and integrates a novel visual feature fusion mechanism to effectively encode high-resolution details and remote sensing category prior of target regions while maintaining global context, leading to highly detailed and context-aware localized captions.

\item We propose a high-quality benchmark, DE-Benchmark, specifically tailored for Geo-DLC. Following the rigorous attribute-based evaluation methodology of DAM, we meticulously design an evaluation dataset that moves beyond traditional reference-based metrics. This ensures models are appropriately rewarded for providing rich, accurate details pertinent to the remote sensing context and penalized for factual errors or irrelevant descriptions, thereby enabling comprehensive and fair assessment.

\end{itemize}

\section{Related Work}

\subsection{Image Captioning}

Image Captioning is a fundamental and critical research problem in the multi-modal field, which has attracted a lot of research due to its complexity. Early works were mainly inspired by natural language processing tasks such as machine translation~\cite{bahdanau2015neural_10}, and adopted attention-based Encoder-Decoder architecture as the mainstream scheme~\cite{vinyals2015show_130, anderson2018bottomup_7, yao2018exploring_149, gu2018stack_47, huang2019attention_65, pan2020xlinear_105}. These methods usually use region-based CNN (\textit{e.g.}, ResNet~\cite{he2016resnet_53}) or Transformer-based backbone~\cite{vaswani2017attention} as the Encoder to extract local region features. Then RNN or LSTM is used as Decoder to generate natural language description. At this stage, a large number of studies focus on multi-scale attention and feature modeling in order to improve the understanding of images. However, limited by the expressive power of convolution operations, the model still has shortcomings in capturing the relationship between high-attention regions.
Subsequently, several works noticed the advantages of graph structure in modeling element relationships and semantic dependencies ~\cite{you2016semantic_150, wu2016value_137, gan2017semantic_36, yao2017boosting_148, zhou2017watch_158} and began to try to construct scene graphs from images~\cite{xu2017scenegraph_139, yang2018graphrcnn_144, zhang2017vte_154, li2017scenegraph_83, tang2019dynamic_127, gu2019scenegraph_50, dai2017drnet_25} or text \cite{wang2018dependency_136, anderson2016spice_6}. It is introduced into the Encoder-Decoder framework \cite{yang2019autoencoding_145} to enhance the semantic alignment of image-text. These methods have made some progress in improving the accuracy of descriptions, but their generalization ability is still limited due to their reliance on small-scale fully supervised training.


With the rise of large-scale pre-trained models~\cite{radford2021learning}, researchers have begun to explore the introduction of pre-training paradigms into image captioning tasks \cite{radford2019language_111, barraco2022clip_12, xia2021xgpt_138}. Typical representatives such as ClipCap\cite{mokady2021clipcap_99} significantly improved the generalization of the model by mapping the embeddings extracted by CLIP into the generative network. Since then, more works have focused on how to better align pre-trained visual features with text generation tasks to further improve the performance. At the same time, there are also studies that try to apply reinforcement learning and unsupervised learning methods to this task \cite{gu2018pivot_48, feng2019unsupervised_33, chen2019cgan_16}, focusing on image feature modeling and cross-modal alignment, but the completeness and length of the generated text are still limited by the Decoder architecture.
In recent years, with the rapid development of MLLMs, the research paradigm of image captioning has further evolved. Some works directly leverage MLLMs to improve the richness and fluency of natural language output. For example, DLC task and DAM enable fine-grained natural language descriptions at the object level~\cite{lian2025describe}, but the performance of DAM is not ideal in remote sensing image scenes, and the detailed language representation at the object level still faces great challenges. In the field of remote sensing, EagleVision introduces object-level attribute-guided formatting descriptions. However, its templated output and limited categories constrain its ability to describe details and its generalizability~\cite{jiang2025eaglevision}. 

\begin{figure*}
    \centering
    \includegraphics[width=1.0\linewidth]{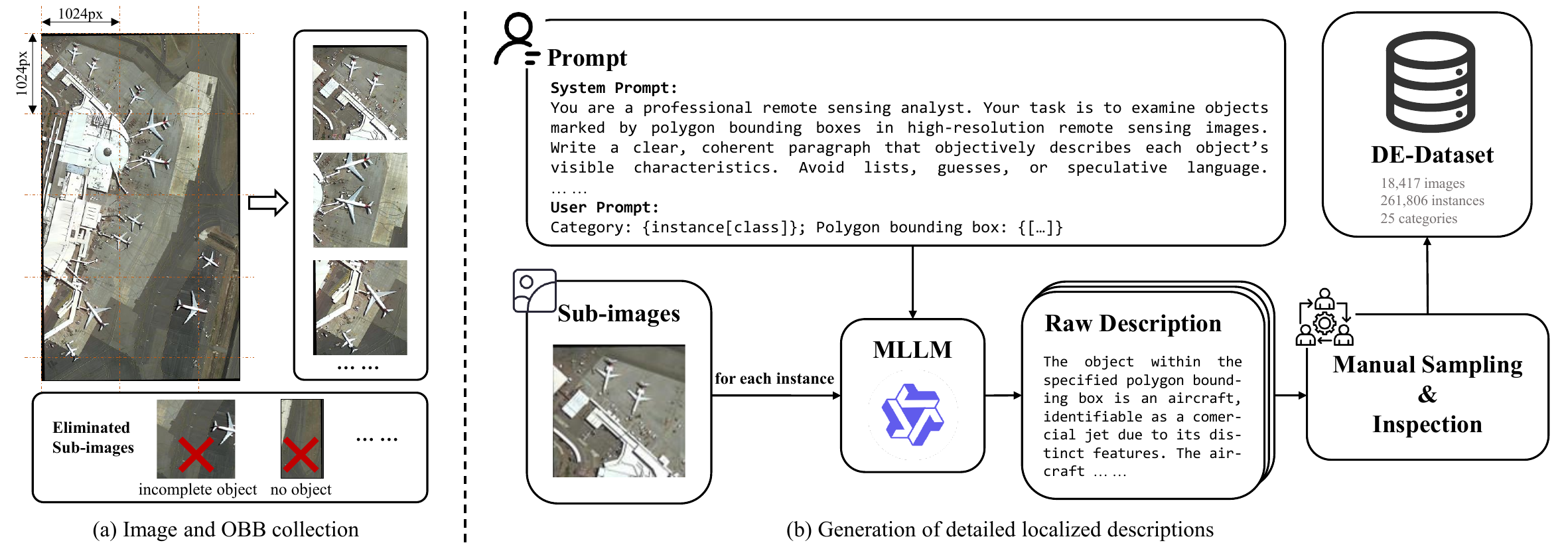}
    \caption{The production pipeline of our DE-Dataset. (a) is the tiling strategy used in dataset preprocessing. (b) is the generation process of localized description.}
    \label{fig:data_pipeline}
\end{figure*}

\subsection{Vision Language Foundation Models for Remote Sensing}

With the rapid advancement of multimodal learning, VLMs have emerged as a critical paradigm for supporting diverse downstream tasks. Remote sensing, as a highly application-driven domain, has also witnessed a growing body of research leveraging VLMs. Within the contrastive learning framework represented by CLIP~\cite{radford2021learning}, numerous studies have sought to adapt the model to remote sensing scenarios. RemoteCLIP~\cite{liu2024remoteclip}, GeoRSCLIP~\cite{GeoRSCLIP} and Git-RSCLIP~\cite{liu2025text2earth} construct large-scale image–text datasets and fine-tune CLIP to enhance performance across multiple tasks. CRSR~\cite{CRSR} and APPLeNet~\cite{singha2023applenet} introduce feature integration modules to strengthen cross-modal alignment, while ProGEO~\cite{2024ProGEO} and GRAFT~\cite{GRAFT} incorporate auxiliary information to improve both image encoding and modality interaction. S-CLIP~\cite{MoKimLeeShin2024SClip} and RS-CLIP~\cite{li2023rs} further refine CLIP’s domain adaptation through improved pseudo-labeling strategies and loss designs. Although these contrastive learning approaches achieve substantial performance gains, their scope remains confined to image–text matching, lacking the capability for natural language generation.

To address this limitation, recent studies have extended remote sensing applications to large-scale conversational VLM architectures. RS-LLaVA~\cite{bazi2024rs} builds upon CLIP as the vision encoder and adapts the LLaVA~\cite{liu2023visual} framework through instruction tuning, enabling dialog-based remote sensing VLMs. EarthGPT~\cite{zhang2024earthgpt} and EarthMarker~\cite{zhang2024earthmarker} optimize vision encoders to enhance multi-scale object representation, while VHM~\cite{pang2025vhm} and SkySenseGPT~\cite{luo2024skysensegpt} mitigate CLIP’s limited spatial awareness by constructing higher-quality image–text pairs. In parallel, several works such as RSGPT~\cite{hu2025rsgpt}, LHRS-Bot~\cite{muhtar2024lhrs}, TEOChat~\cite{irvin2024teochat}, and UniRS~\cite{li2024unirs} explore Vision–LLM connector architectures to improve multimodal perception. Beyond architectural advances, some methods have been proposed to address specific challenges of remote sensing VLMs. For instance, chain-of-thought reasoning and data augmentation strategies have been introduced to improve reasoning and generalization capabilities, as demonstrated in CPSeg~\cite{li2024cpseg}, MGeo~\cite{ding2023mgeo}, and SpectralGPT~\cite{hong2023spectralgpt}.

\subsection{Remote Sensing Datasets}

Currently, most existing remote sensing VLMs are data-driven, and in general, different tasks rely on their specific datasets. For remote sensing visual question answering, EarthVQA~\cite{wang2024earthvqa} and CRSVQA~\cite{zhang2023multistep} adopt manual annotation to ensure high-quality supervision. For object detection, datasets such as FAIR1M~\cite{sun2022fair1m}, DOTA~\cite{xia2018dota}, and DIOR~\cite{li2020object} have been developed, while classification tasks are supported by AID~\cite{xia2017aid}, NWPU-RESISC45~\cite{cheng2017remote}, and UCM~\cite{yang2010bag}, and semantic segmentation relies on pixel-level datasets such as iSAID~\cite{waqas2019isaid}. Building on these resources, several extended datasets have been proposed by combining or re-annotating existing ones to address new challenges, including RSVGD~\cite{zhan2023rsvg}, RefSegRS~\cite{yuan2023rrsis}, and SkyEye-968K~\cite{zhan2025skyeyegpt}. Although these datasets cover fundamental tasks and some emerging applications, the demand for large-scale remote sensing vision-language datasets is growing rapidly with the development of MLLMs. To address this, researchers have explored automated annotation pipelines supported by general foundation models. GeoChat~\cite{kuckreja2024geochat} designs system-level prompts to interact with Vicuna~\cite{chiang2023vicuna} for generating multi-turn question–answer pairs. HqDC-1.4M~\cite{pang2025vhm} employs Gemini-1.0-pro-vision to produce textual descriptions for large-scale remote sensing images, yielding extensive image–text pairs. FIT-RS~\cite{luo2024skysensegpt} leverages TinyLLaVA-3.1B~\cite{zhou2024tinyllava} to generate background descriptions of remote sensing images and integrates GPT-4 or GPT-3.5 for high-quality annotation. Despite these advances, high-quality fine-grained object-level datasets remain scarce, and current pipelines for generating object-level descriptions are still immature.

\section{Task, Dataset and Benchmark}

This section formally defines the Geo-DLC task, details the construction of DE-Dataset, and introduces the DE-Benchmark. These address the challenges in Section~\ref{sec:intro}.

\subsection{Task Formulation}


Following the framework of DLC as established in~\cite{lian2025describe}, the Geo-DLC task extends this concept to the remote sensing image. Unlike general image captioning which provides a holistic summary, Geo-DLC focuses on generating precise, comprehensive textual descriptions for specific geographical regions or objects within remote sensing images. Formally, given an input image $I$ and a user-specified geographical Region of Interest (ROI) $R$ within $I$ (typically represented by various localization cues such as a bounding box, point, or binary mask), the objective is to generate a detailed textual description $T$ centered on the instance. Such descriptions are required to cover both intrinsic object features and contextual environmental attributes, ensuring logical consistency and informativeness. This task can be formulated as:

 


\begin{equation}
\begin{aligned}
T = \textit{Model}(I, R)
\label{eq:task_define}
\end{aligned}
\end{equation}

The Geo-DLC task is challenging due to the inherent differences in object appearance from an aerial viewpoint, the vast scale variations, and the complex background contexts prevalent in remote sensing images. It demands a model to not only recognize objects under these unique conditions but also to elaborate on their fine-grained attributes and relationships, which existing general MLLMs often fail to achieve.

\subsection{DE-Dataset}

To support Geo-DLC task, we introduce DE-Dataset, the first Geo-DLC dataset constructed through a multi-stage process leveraging existing remote sensing datasets, MLLMs for automated annotation, and rigorous human quality control.

\subsubsection{Data Construction Pipeline}

We construct our DE-Dataset based on two oriented object detection datasets: DIOR~\cite{li2020object} and DOTA~\cite{xia2018dota}. 
To manage the large variation in image scales within DOTA and ensure consistent input for annotation, we perform a unified tiling strategy.
Specifically, for large-scale images with spatial resolutions greater than $1024 \times 1024$, we partition them into non-overlapping $1024 \times 1024$ sub-images, as well as additional boundary tiles, to preserve the diversity of instance placements. 
All instance bounding boxes are then updated relative to the sub-image coordinates, and some low-quality sub-images are eliminated (\textit{e.g.}, those containing only incomplete objects or no any objects, etc.). This preprocessing yields a refined collection of remote sensing sub-images and their corresponding OBBs, as shown in Fig. \ref{fig:data_pipeline}~(a). 






\begin{figure}[t]
    \centering
    \includegraphics[width=1\linewidth]{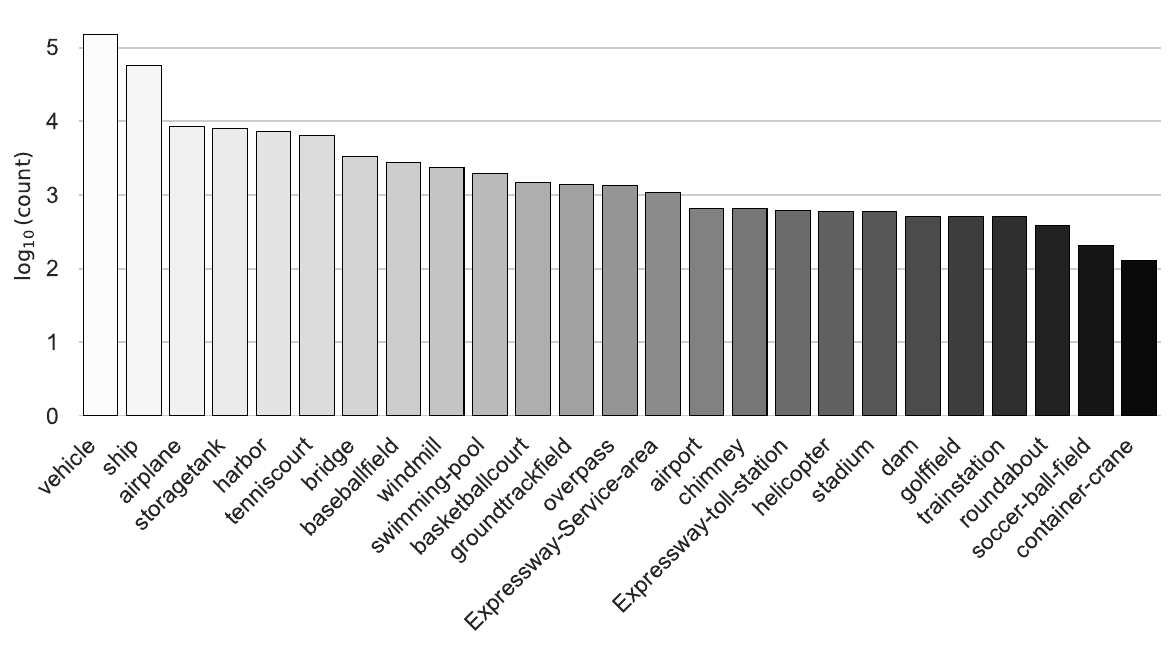}
    \caption{The distribution of instance categories and their corresponding quantities ($log_{10}$) in DE-Dataset dataset.}
    \label{fig:class_dist}
\end{figure}

To generate detailed localized captions at scale, we utilize a powerful MLLM, Qwen2.5-VL-32B~\cite{bai2025qwen2}, as an automated annotator. To alleviate cognitive load on the annotator and produce more accurate descriptions, we provide corresponding class name, sub-image, and its OBB for each object instance as input. We design prompts explicitly instructing the model to generate descriptions covering the following key aspects:

\begin{itemize}
    \item Intrinsic features: appearance, structure, and other object-specific characteristics.

    \item Spatial features: orientation, position, and other localization attributes.

    \item Contextual features: surrounding environment, as well as signs of human or social activity.
\end{itemize}

In addition, to ensure dataset quality and reduce hallucinations, we impose strict linguistic constraints on the model output. Speculative or uncertain language is explicitly prohibited, ensuring that the generated descriptions remain logical, consistent, and factually grounded. Our prompt is shown in Fig.~\ref{fig:data_pipeline}~(b).
Furthermore, after the initial dataset is obtained, a random subset of the automatically generated captions, covering a diverse range of categories and complexities, is meticulously reviewed by human experts with remote sensing domain knowledge. These experts verify the factual correctness, descriptive richness, and adherence to the Geo-DLC task definition. Incorrect or low-quality captions are either revised, regenerated, or discarded. The overall construction pipeline is illustrated in Fig.~\ref{fig:data_pipeline}. 





\subsubsection{Dataset Statistics}

\begin{figure}[t]
    \centering
    \includegraphics[width=\linewidth]{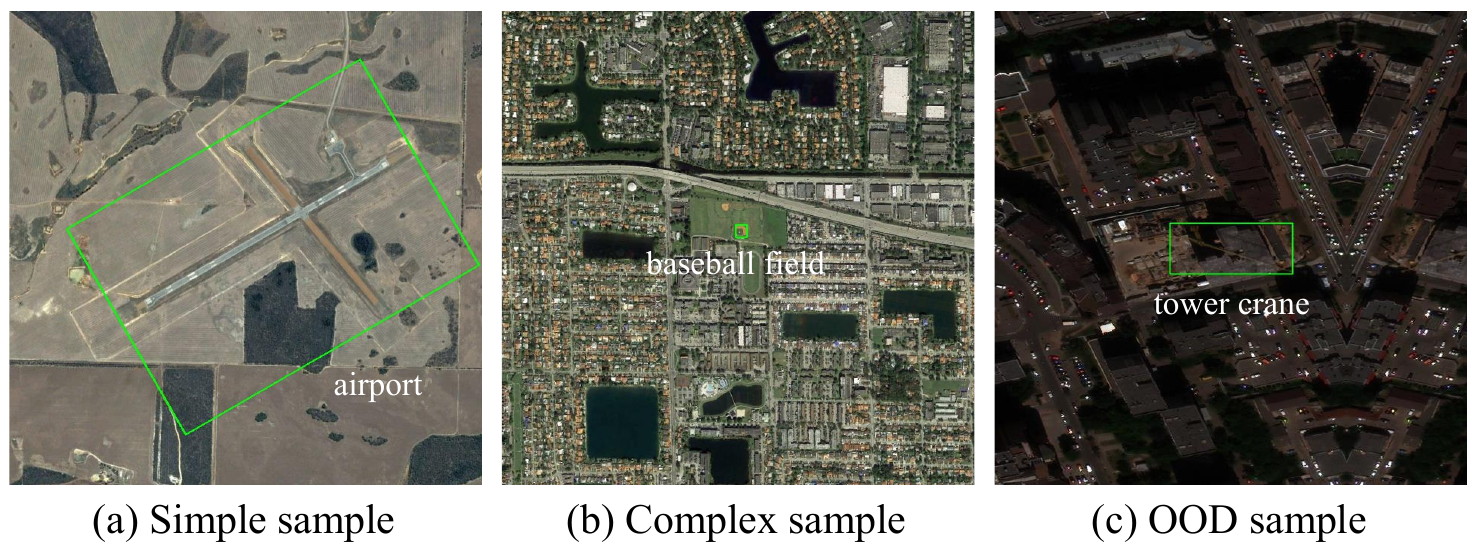}
    \caption{Examples of different instance types in DE-Benchmark.}
    \label{fig:different}
\end{figure}

The DE-Dataset comprises 18,417 images and 261,806 instances, with an average of 14 instances per image. Each instance is accompanied by a detailed description averaging 119.83 words in length.
The DE-Dataset contains 25 categories, with the category distribution shown in Fig.~\ref{fig:class_dist}.
Unlike traditional remote sensing image captioning datasets that offer only coarse-grained annotations or full-image descriptions~\cite{qu2016deep, lu2017exploring, cheng2022nwpu, liu2022remote} and in contrast to object-level attribute datasets like EVAttrs-95K~\cite{jiang2025eaglevision} which provide templated descriptions for a limited number of categories, DE-Dataset stands out. By adopting an extended paragraph form for descriptions, DE-Dataset offers a richer, more open-ended, and flexible representation of instances, encompassing both intrinsic attributes and contextual information without predefined templates. This design significantly enhances the practicality and versatility of the dataset for Geo-DLC tasks.


\subsection{DE-Benchmark}



To provide a robust and fair evaluation for Geo-DLC models, we construct DE-Benchmark, which adapts and extends the attribute-based evaluation protocol established in~\cite{xiao2025describe}. This protocol moves beyond traditional reference-based metrics by assessing factual accuracy and descriptive richness against a set of predefined attributes, a critical feature for Geo-DLC given the often incomplete nature of reference captions.


\subsubsection{Preliminary Data Collection}

Following the procedure used in DE-Dataset, we construct our benchmark based on the validation sets of DIOR and DOTA. Reusing the proposed data construction pipeline, we first generate detailed descriptions of the instances. For each category, several representative instances are manually selected to ensure coverage, resulting in preliminary evaluation samples. Considering the significant variations among remote sensing images across different regions, as well as issues such as dense distribution, large-scale changes, occlusion, and complex contextual backgrounds of some remote sensing objects, we further classify instances into simple and complex samples based on instance size, spatial location, occlusion level, contextual complexity, etc. Furthermore, to evaluate the model's generalization and open-set recognition ability, we additionally selected 10 out-of-distribution (OOD) categories from the xView~\cite{lam2018xview} dataset which are not included in the DE-Dataset. Their descriptions are generated using the same pipeline, and the resulting samples are merged with the preliminary benchmark data. Some examples showing the different types of instances are shown in Fig.~\ref{fig:different}.





\subsubsection{Attribute-Based QA Construction}

\begin{figure*}
    \centering
    \includegraphics[width=\textwidth]{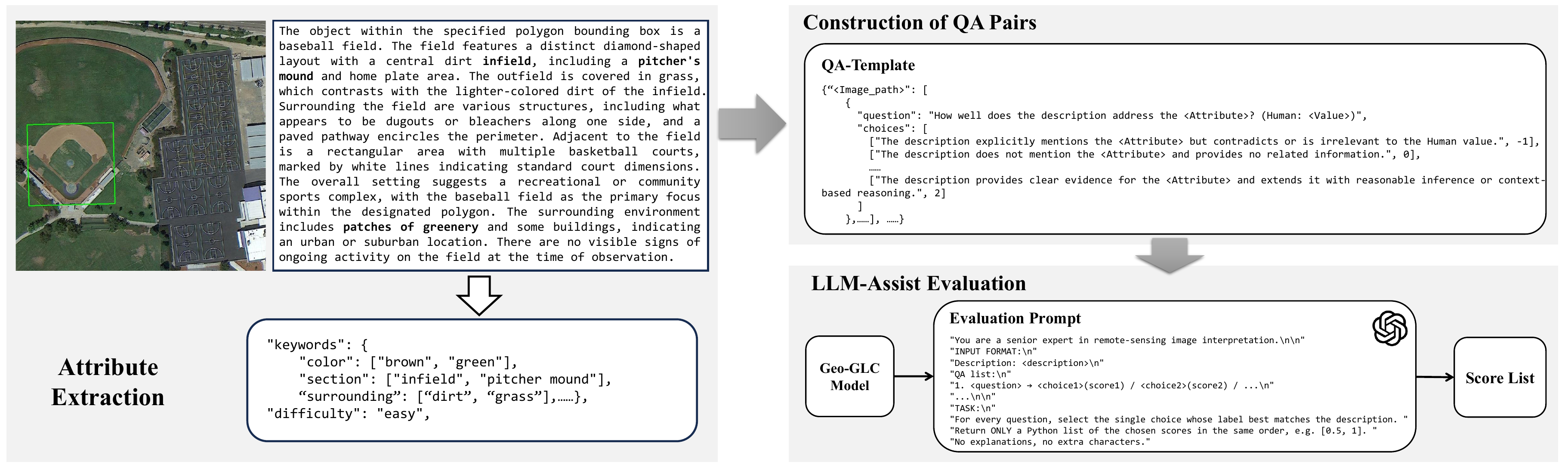}
    \caption{The workflow of construction and evaluation for DE-Benchmark. We first manually  extract key attributes and define difficulty for instances, which are then used to generate structured QA pairs with predefined scoring rubrics. Finally, an LLM evaluates the Geo-DLC model's generated descriptions against these QA pairs to produce a score.}
    \label{fig:benchmark_workflow}
\end{figure*}

For every instance in the DE-Benchmark, human experts carefully extract a set of attribute-value pairs. An attribute specifies a key feature that should be present in the description, \textit{e.g.}, color, neighboring objects, spatial orientation. The corresponding value is the accurate, human-annotated information for that attribute. This manual annotation provides the ground truth for evaluating the factual accuracy and quality of model-generated captions, as shown on the left side of Fig.~\ref{fig:benchmark_workflow}.
Based on these attribute-value pairs, we create QA templates for evaluation. Each attribute leads to a question (\textit{e.g.}, "How well does the description address the color?"). For each question, we design a set of detailed options. These options are scoring rubrics, not factual answers. They describe various levels of how well the model's output covers or presents the specific attribute. These rubrics include whether the description contradicts the human value, does not mention it, gives vague hints, supports it through implication, provides clear evidence, or extends it with reasonable inference. Each rubric option has a pre-assigned numerical score, allowing for a fine-grained assessment of the model's descriptive quality for each attribute.


\subsubsection{LLM-Assist Evaluation}

Evaluating detailed, open-ended descriptions in Geo-DLC is challenging. For this, we develop an LLM-Assist evaluation system that uses an LLM as a judge. Specifically, we adopt GPT-4.1, given its strong reasoning and language understanding.
For each evaluation instance, the Geo-DLC model first generates its detailed localized caption. Then, this caption, along with its corresponding attribute-based QA pairs, is provided as input to the judgment LLM. The LLM is prompted to analyzes the model's description and, for each question, selects the rubric option that best matches how the attribute is addressed in the generated caption. Moreover, 5 general language quality questions further evaluate the description's grammatical correctness, logical flow, factual reliability, inferential capability, and conciseness. For structured analysis, all questions are categorized into four main types: ``Appearance'' (intrinsic visual characteristics), ``Surrounding'' (environmental and contextual information), ``Language'' (grammatical correctness, logical flow, and naturalness of descriptions), and ``Usage'' (functional properties and potential usage scenarios). The construction and evaluation process of DE-Benchmark is shown in Fig.~\ref{fig:benchmark_workflow}.

\section{Methodology}

\begin{figure*}
    \centering
    \includegraphics[width=\textwidth]{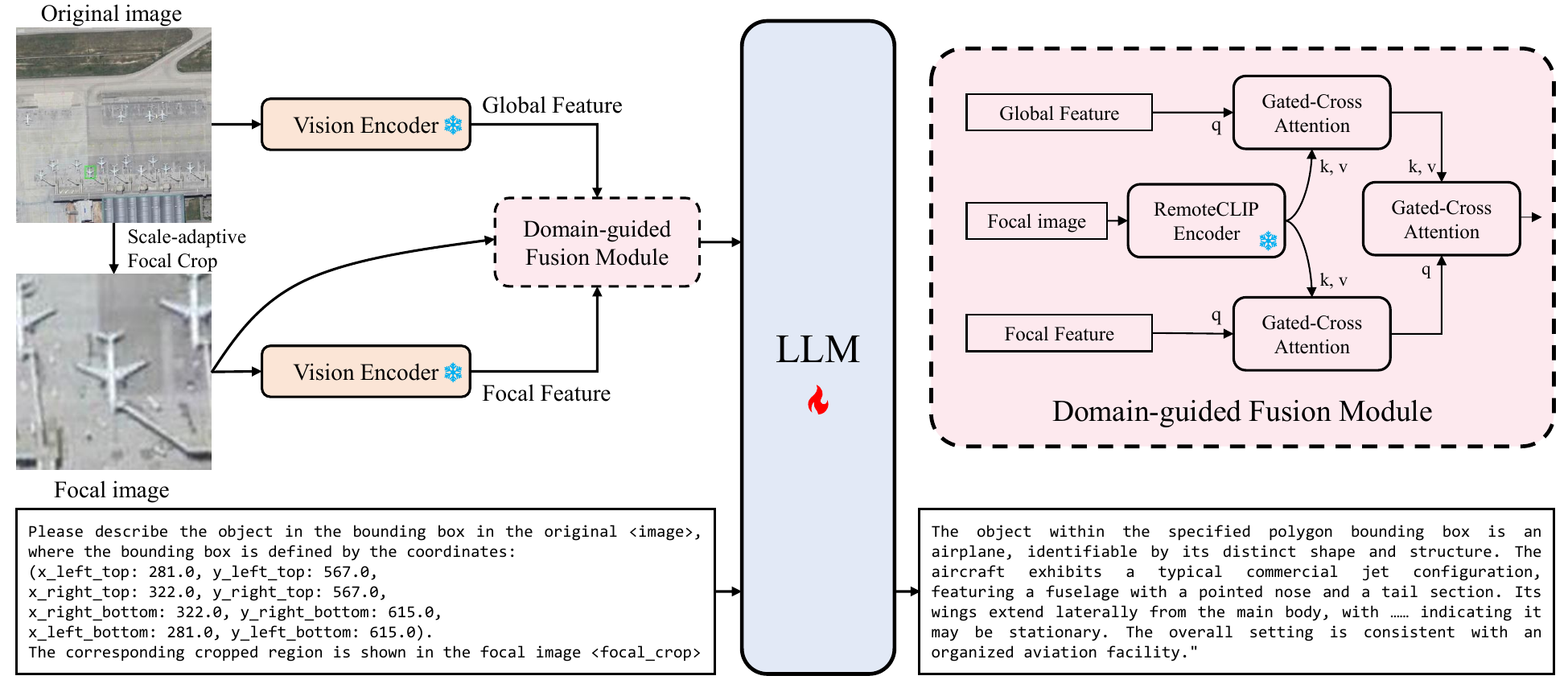}
    \caption{The architecture of DescribeEarth. It uses a scale-adaptive focal strategy to extract both global and focal visual features from an original image. These features are then integrated with domain-specific priors from a RemoteCLIP encoder via DFM. The enriched visual representation, along with object bounding box coordinates embedded in the textual prompt, is then fed to the LLM for generating detailed localized captions.}
    \label{fig:overview}
\end{figure*}

To address the challenges of Geo-DLC mentioned above, \textit{i.e.}, large-scale variations, limited recognition capabilities from the remote sensing perspective, and the lack of detailed and contextual descriptions, we propose DescribeEarth. Built upon the Qwen2.5-VL-3B-Instruct, DescribeEarth incorporates a scale-adaptive focal strategy to manage diverse object scales and a domain-guided fusion module (DFM) to integrate domain-specific semantic priors. These modules enhance the model's ability to generate highly detailed and accurate localized captions for remote sensing images. An overview of the DescribeEarth architecture is illustrated in Fig.~\ref{fig:overview}.


\subsection{Scale-adaptive Focal Strategy}

Inspired by DAM's Focal Prompt strategy~\cite{lian2025describe}, which balances local detail with global context in natural images, we develop a scale-adaptive focal strategy for Geo-DLC. Remote sensing images, with their nadir views and extreme variations in object scales—from expansive aircraft runways to minute vehicles—present unique challenges for direct application of fixed-size focal crops. Our strategy improves DAM by dynamically generating a focal crop that, alongside the full original global image, ensures high-resolution encoding of the target region while maintaining crucial surrounding context, tailored for remote sensing characteristics. Here, the scale of each object is precisely determined by the longer side of its minimum enclosing axis-aligned bounding box. Based on this definition, we implement three distinct mechanisms for generating the focal crop: 



\begin{itemize}
    \item Large objects ($>224$ pixel): For objects whose original bounding box is already large enough to encompass sufficient fine-grained detail, the bounding box itself is directly cropped to form the focal region.
    \item Medium objects ($112$-$224$ pixel): A fixed $224 \times 224$ patch, centered at the bounding box centroid, is extracted as the focal crop. This patch size is designed to capture both fine object details and relevant surrounding context.
    \item Small objects ($<112$ pixel): A $112 \times 112$ square patch is extracted around the bounding box center. This tighter crop ensures that the limited pixels available for the target object receive maximum attention, emphasizing its details while retaining sufficient environmental cues to aid recognition and description.
\end{itemize}


All these generated focal crops, irrespective of their original scale, are then resized to a uniform resolution to match the input size of the vision encoder. This focal crop, alongside the original global image, then serves as input to our dual-path vision encoding pipeline, ensuring comprehensive feature extraction that leverages both fine-grained local details and broader contextual signals, as shown in Fig.~\ref{fig:overview}.

\begin{table*}[ht]
\centering
\caption{Performance comparison across categories, difficulty levels, and question types (in \%). The numbers in parentheses indicate the gap between our DescribeEarth model and other models. Positive values in green, negative in red. ``ft'' indicates fine-turning on our DE-Dataset.}
\label{tab:main_full}
\scalebox{0.95}{
\begin{tabular}{llccccccc}
\hline
& Key & DescribeEarth & Qwen2.5-VL-3B (ft) & Qwen2.5-VL-3B & GPT-4o & Gemini-2.5-Pro & DAM \\
\hline
Category & Baseball field & \textbf{75.75} & 62.00 (\textcolor{green}{+13.75}) & 66.00 (\textcolor{green}{+9.75}) & 56.00 (\textcolor{green}{+19.75}) & 71.25 (\textcolor{green}{+4.50}) & 72.00 (\textcolor{green}{+3.75}) \\
& Tennis court & \textbf{78.25} & 77.50 (\textcolor{green}{+0.75}) & 61.00 (\textcolor{green}{+17.25}) & 63.00 (\textcolor{green}{+15.25}) & 60.00 (\textcolor{green}{+18.25}) & 56.50 (\textcolor{green}{+21.75}) \\
& Stadium & 75.50 & \textbf{76.50} (\textcolor{red}{-1.00}) & 76.00 (\textcolor{red}{-0.50}) & 69.00 (\textcolor{green}{+6.50}) & 51.25 (\textcolor{green}{+24.25}) & 59.00 (\textcolor{green}{+16.50}) \\
& Harbor & \textbf{86.34} & 84.79 (\textcolor{green}{+1.55}) & 72.16 (\textcolor{green}{+14.18}) & 73.45 (\textcolor{green}{+12.89}) & 60.57 (\textcolor{green}{+25.77}) & 60.82 (\textcolor{green}{+25.52}) \\
& Golf field & \textbf{68.85} & 65.98 (\textcolor{green}{+2.87}) & 59.22 (\textcolor{green}{+9.63}) & 67.42 (\textcolor{green}{+1.43}) & 46.11 (\textcolor{green}{+22.74}) & 60.86 (\textcolor{green}{+7.99}) \\
& Ground track field & \textbf{83.43} & 77.53 (\textcolor{green}{+5.90}) & 74.44 (\textcolor{green}{+8.99}) & 74.44 (\textcolor{green}{+8.99}) & 70.79 (\textcolor{green}{+12.64}) & 64.89 (\textcolor{green}{+18.54}) \\
& Airport & \textbf{74.31} & 71.30 (\textcolor{green}{+3.01}) & 61.81 (\textcolor{green}{+12.50}) & 71.30 (\textcolor{green}{+3.01}) & 42.59 (\textcolor{green}{+31.72}) & 61.34 (\textcolor{green}{+12.97}) \\
& Expressway service area & 56.40 & 58.53 (\textcolor{red}{-2.13}) & 55.62 (\textcolor{green}{+0.78}) & \textbf{61.63} (\textcolor{red}{-5.23}) & 37.40 (\textcolor{green}{+19.00}) & 51.74 (\textcolor{green}{+4.66}) \\
& Overpass & 68.81 & \textbf{72.14} (\textcolor{red}{-3.33}) & 68.10 (\textcolor{green}{+0.71}) & 65.48 (\textcolor{green}{+3.33}) & 60.24 (\textcolor{green}{+8.57}) & 56.43 (\textcolor{green}{+12.38}) \\
& Expressway toll station & 65.74 & 65.97 (\textcolor{red}{-0.23}) & 53.47 (\textcolor{green}{+12.27}) & \textbf{67.59} (\textcolor{red}{-1.85}) & 41.90 (\textcolor{green}{+23.84}) & 53.01 (\textcolor{green}{+12.73}) \\
& Airplane & 78.80 & \textbf{80.00} (\textcolor{red}{-1.20}) & 64.67 (\textcolor{green}{+14.13}) & 75.00 (\textcolor{green}{+3.80}) & 48.37 (\textcolor{green}{+30.43}) & 70.92 (\textcolor{green}{+7.88}) \\
& Storage tank & 80.00 & \textbf{81.50} (\textcolor{red}{-1.50}) & 75.25 (\textcolor{green}{+4.75}) & 73.00 (\textcolor{green}{+7.00}) & 65.00 (\textcolor{green}{+15.00}) & 63.75 (\textcolor{green}{+16.25}) \\
& Train station & \textbf{83.78} & 79.52 (\textcolor{green}{+4.26}) & 68.88 (\textcolor{green}{+14.90}) & 73.94 (\textcolor{green}{+9.84}) & 57.18 (\textcolor{green}{+26.60}) & 62.23 (\textcolor{green}{+21.55}) \\
& Ship & \textbf{83.33} & 83.33 (\textcolor{green}{0.00}) & 65.44 (\textcolor{green}{+17.89}) & 73.04 (\textcolor{green}{+10.29}) & 56.37 (\textcolor{green}{+26.96}) & 65.69 (\textcolor{green}{+17.64}) \\
& Windmill & 71.24 & \textbf{72.04} (\textcolor{red}{-0.80}) & 69.62 (\textcolor{green}{+1.62}) & 67.20 (\textcolor{green}{+4.04}) & 61.02 (\textcolor{green}{+10.22}) & 60.48 (\textcolor{green}{+10.76}) \\
& Basketball court & 53.64 & 56.82 (\textcolor{red}{-3.18}) & 50.00 (\textcolor{green}{+3.64}) & \textbf{65.45} (\textcolor{red}{-11.81}) & 39.09 (\textcolor{green}{+14.55}) & 50.68 (\textcolor{green}{+2.96}) \\
& Chimney & 63.29 & 63.06 (\textcolor{green}{+0.23}) & 60.59 (\textcolor{green}{+2.70}) & \textbf{66.44} (\textcolor{red}{-3.15}) & 38.29 (\textcolor{green}{+25.00}) & 59.01 (\textcolor{green}{+4.28}) \\
& Dam & \textbf{73.41} & 72.73 (\textcolor{green}{+0.68}) & 69.77 (\textcolor{green}{+3.64}) & 72.50 (\textcolor{green}{+0.91}) & 38.18 (\textcolor{green}{+35.23}) & 64.77 (\textcolor{green}{+8.64}) \\
& Bridge & 68.87 & 68.63 (\textcolor{green}{+0.24}) & 68.14 (\textcolor{green}{+0.73}) & \textbf{71.57} (\textcolor{red}{-2.70}) & 42.65 (\textcolor{green}{+26.22}) & 63.73 (\textcolor{green}{+5.14}) \\
& Building & 67.71 & 65.10 (\textcolor{green}{+2.61}) & \textbf{70.31} (\textcolor{red}{-2.60}) & 61.46 (\textcolor{green}{+6.25}) & 41.67 (\textcolor{green}{+26.04}) & 16.67 (\textcolor{green}{+51.04}) \\
& Construction site & 65.82 & 54.59 (\textcolor{green}{+11.23}) & 65.31 (\textcolor{green}{+0.51}) & \textbf{79.08} (\textcolor{red}{-13.26}) & 43.88 (\textcolor{green}{+21.94}) & 55.10 (\textcolor{green}{+10.72}) \\
& Damaged building & 56.63 & 55.10 (\textcolor{green}{+1.53}) & 53.57 (\textcolor{green}{+3.06}) & \textbf{62.24} (\textcolor{red}{-5.61}) & 40.82 (\textcolor{green}{+15.81}) & 21.43 (\textcolor{green}{+35.20}) \\
& Facility & \textbf{88.37} & 68.02 (\textcolor{green}{+20.35}) & 83.14 (\textcolor{green}{+5.23}) & 72.67 (\textcolor{green}{+15.70}) & 46.51 (\textcolor{green}{+41.86}) & 41.86 (\textcolor{green}{+46.51}) \\
& Locomotive & \textbf{72.22} & 61.11 (\textcolor{green}{+11.11}) & 56.67 (\textcolor{green}{+15.55}) & 63.33 (\textcolor{green}{+8.89}) & 45.56 (\textcolor{green}{+26.66}) & 21.67 (\textcolor{green}{+50.55}) \\
& Roundabout & \textbf{69.89} & 69.35 (\textcolor{green}{+0.54}) & 64.52 (\textcolor{green}{+5.37}) & 64.78 (\textcolor{green}{+5.11}) & 50.81 (\textcolor{green}{+19.08}) & 54.57 (\textcolor{green}{+15.32}) \\
& Helicopter & 67.38 & \textbf{70.24} (\textcolor{red}{-2.86}) & 62.38 (\textcolor{green}{+5.00}) & 60.71 (\textcolor{green}{+6.67}) & 39.52 (\textcolor{green}{+27.86}) & 60.71 (\textcolor{green}{+6.67}) \\
& Large vehicle & \textbf{73.16} & 70.79 (\textcolor{green}{+2.37}) & 59.74 (\textcolor{green}{+13.42}) & 63.68 (\textcolor{green}{+9.48}) & 34.74 (\textcolor{green}{+38.42}) & 59.74 (\textcolor{green}{+13.42}) \\
& Pylon & 67.07 & 57.32 (\textcolor{green}{+9.75}) & 60.37 (\textcolor{green}{+6.70}) & \textbf{70.12} (\textcolor{red}{-3.05}) & 48.78 (\textcolor{green}{+18.29}) & 21.34 (\textcolor{green}{+45.73}) \\
& Shed & 62.24 & 62.76 (\textcolor{red}{-0.52}) & 62.24 (\textcolor{green}{0.00}) & \textbf{67.86} (\textcolor{red}{-5.62}) & 40.82 (\textcolor{green}{+21.42}) & 44.90 (\textcolor{green}{+17.34}) \\
& Tower & 66.28 & 64.53 (\textcolor{green}{+1.75}) & 66.28 (\textcolor{green}{0.00}) & \textbf{71.51} (\textcolor{red}{-5.23}) & 46.51 (\textcolor{green}{+19.77}) & 20.93 (\textcolor{green}{+45.35}) \\
& Tower crane & 64.58 & 63.54 (\textcolor{green}{+1.04}) & 54.17 (\textcolor{green}{+10.41}) & \textbf{68.75} (\textcolor{red}{-4.17}) & 33.33 (\textcolor{green}{+31.25}) & 0.00 (\textcolor{green}{+64.58}) \\
& Vehicle lot & 80.23 & \textbf{87.79} (\textcolor{red}{-7.56}) & 65.70 (\textcolor{green}{+14.53}) & 71.51 (\textcolor{green}{+8.72}) & 44.19 (\textcolor{green}{+36.04}) & 37.79 (\textcolor{green}{+42.44}) \\
\hline
Difficulty Level & Simple & \textbf{74.19} & 68.07 (\textcolor{green}{+6.12}) & 66.81 (\textcolor{green}{+7.38}) & 70.24 (\textcolor{green}{+3.95}) & 53.30 (\textcolor{green}{+20.89}) & 62.20 (\textcolor{green}{+11.99}) \\
& Complex & \textbf{70.24} & 63.76 (\textcolor{green}{+6.48}) & 62.19 (\textcolor{green}{+8.05}) & 65.51 (\textcolor{green}{+4.73}) & 46.77 (\textcolor{green}{+23.47}) & 58.54 (\textcolor{green}{+11.70}) \\
& OOD & \textbf{68.78} & 63.75 (\textcolor{green}{+5.03}) & 63.59 (\textcolor{green}{+5.19}) & 68.78 (\textcolor{green}{0.00}) & 43.01 (\textcolor{green}{+25.77}) & 28.22 (\textcolor{green}{+40.56}) \\
\hline
Question Type & Appearance & 42.26 & \textbf{53.77} (\textcolor{red}{-11.51}) & 31.43 (\textcolor{green}{+10.83}) & 35.49 (\textcolor{green}{+6.77}) & 13.94 (\textcolor{green}{+28.32}) & 31.94 (\textcolor{green}{+10.32}) \\
& Surround & \textbf{54.41} & 37.86 (\textcolor{green}{+16.55}) & 35.68 (\textcolor{green}{+18.73}) & 43.99 (\textcolor{green}{+10.42}) & 11.32 (\textcolor{green}{+43.09}) & 6.46 (\textcolor{green}{+47.95}) \\
& Language & \textbf{98.77} & 98.07 (\textcolor{green}{+0.70}) & 96.88 (\textcolor{green}{+1.89}) & 98.14 (\textcolor{green}{+0.63}) & 85.43 (\textcolor{green}{+13.34}) & 86.88 (\textcolor{green}{+11.89}) \\
& Usage & 35.49 & 32.37 (\textcolor{green}{+3.12}) & 31.03 (\textcolor{green}{+4.46}) & \textbf{44.64} (\textcolor{red}{-9.15}) & 13.62 (\textcolor{green}{+21.87}) & 20.98 (\textcolor{green}{+14.51}) \\
\hline
\end{tabular}}
\end{table*}

\subsection{Vision Encoding and Domain-guided Fusion}

Both the global image and the focal cropped image are processed by a shared vision encoder, yielding global features and focal features. This encoder is initialized with the visual weights from the pre-trained Qwen2.5-VL-3B model and is kept frozen during training to leverage its general visual understanding. Weight sharing enforces representational consistency and improves parameter efficiency, enabling the model to align cross-scale representations in a unified feature space.


To further enhance semantic richness, we introduce DFM. This module integrates domain-specific priors from RemoteCLIP~\cite{liu2024remoteclip}, which is adept at understanding remote sensing visuals. Specifically, focal crops are also encoded by a frozen RemoteCLIP encoder and then projected via a linear layer to obtain the domain-specific feature. The fusion of these diverse visual cues (global, focal, and domain-specific) is managed through a hierarchical gated cross-attention mechanism~\cite{alayrac2022flamingo}, as shown in Fig.~\ref{fig:overview}. First, global and focal features interact with the domain-specific features in parallel gated cross-attention modules. This fusion aims to semantically enrich both the global and focal representations with remote sensing-specific knowledge. Following this, the enhanced global features are used to refine the focal features in a subsequent cross-attention operation. This operation allows the fine-grained, localized information from the focal features to be further enriched and contextualized by the broader global representation. This ensures that the final visual representation, which is passed to the LLM, is both highly detailed for the localized region and deeply grounded in its wider geographical context.


\subsection{Integration with the LLM}

For integrating localization with the LLM in Geo-DLC, DescribeEarth adopts a strategy distinct from DAM, which directly encodes mask information within its vision backbone via dedicated mask embedding layers.
Recognizing that direct manipulation of visual input with masks can alter the vision encoder's input shape or distribution, potentially requiring extensive fine-tuning or impacting its general understanding, we instead convey the ROI by embedding OBB coordinates directly into the LLM's textual prompt. This design preserves the visual input stream to our shared vision encoder in its pristine, standard image patch format, allowing the powerful, pre-trained vision encoder (from Qwen2.5-VL-3B) to remain frozen and untouched. This avoids new visual biases and ensures robust leveraging of its established general representations. Furthermore, embedding OBB coordinates as text provides the LLM with explicit, numerical spatial information, enabling it to directly reason about the precise location, extent, and crucial orientation of remote sensing objects, which is often paramount for geometrically accurate descriptions. OBBs are inherently superior to horizontal bounding boxes for delineating arbitrarily oriented structures prevalent in nadir-view image, and their textual representation allows the LLM to comprehend these complex geometries without visual perturbation. This text-based localization input is also highly flexible, naturally compatible with various text-conditioned tasks and user interaction paradigms, paving the way for easier integration with future multi-modal applications without altering the visual processing pipeline.


During inference, user interaction for specifying the ROI is highly flexible. Common inputs such as clicks or interactive bounding boxes are first pre-processed into a unified OBB representation. This conversion leverages segmentation models like SAM~\cite{kirillov2023segment}, ensuring that even imprecise user inputs can be transformed into a standardized, precise OBB format. This processed OBB, along with the visual tokens, forms the textual prompt which is then fed to the LLM. The LLM then performs autoregressive decoding, leveraging both these rich visual features and the explicit OBB textual information, to generate the fine-grained, detailed Geo-DLC caption.

\section{Experiments}


\subsection{Experimental Setup}

The DescribeEarth model was trained using two A800 GPUs. The basic architecture is built on Qwen2.5-VL-3B. The vision encoder component of DescribeEarth is initialized from the pre-trained Qwen2.5-VL-3B model and is kept frozen throughout the training process to preserve its established general visual understanding capabilities. To incorporate domain-specific knowledge, RemoteCLIP, a pre-trained ViT-B architecture, is utilized; its weights are also kept frozen, and intermediate outputs are employed rather than only final encoder features to enrich the visual representation. The LLM and the vision-language connector (MLP) within DescribeEarth are initialized with pre-trained Qwen2.5-VL-3B weights. Training was conducted with a batch size of 4, employing gradient accumulation every 4 steps, and a single epoch was run under a fine-finetuning strategy. The global image resolution is set to $448 \times 448$, while the focal image resolution is set to $224 \times 224$.


\subsection{Main Results}

Table~\ref{tab:main_full} presents a detailed performance comparison of DescribeEarth against several state-of-the-art closed-source multimodal large language models (GPT-4o, Gemini-2.5-pro), a DLC model (DAM), and the original Qwen2.5-VL-3B-Instruct on the DE-Benchmark. Overall, DescribeEarth consistently achieves strong performance across various categories and difficulty levels. It outperforms the best closed-source model by a notable margin of 3.95\% on simple instances and 4.73\% on complex instances. While GPT-4o shows competitive performance on some categories (\textit{e.g.}, it is slightly better on ``Basketball court'' and ``Construction site'' categories, and achieves 68.78\%, matching DescribeEarth's score on OOD samples, indicating its broad generalization ability), DescribeEarth significantly surpasses all other compared models, demonstrating its specialized effectiveness for remote sensing images. Specifically, DescribeEarth shows substantial gains in categories like ``Harbor'', ``Ground track field'', ``Facility'', and ``Locomotive'', which are highly relevant and frequent objects in remote sensing. DAM, designed for natural images, struggles significantly on remote sensing data, with gaps as large as +64.58\% relative to DescribeEarth for ``Tower crane'', highlighting the domain adaptation challenge.

When analyzing performance across different question types, DescribeEarth shows particular strengths in ``Surrounding'' descriptions, with a remarkable gain of 10.42\% over GPT-4o and 43.09\% over Gemini-2.5-pro. This indicates the effectiveness of our design in integrating contextual environmental attributes. Furthermore, it exhibits strong performance in ``Appearance'' and ``Language'' type questions, with respective gains of +6.77\% and +0.63\% over GPT-4o. These results underscore DescribeEarth's ability to generate rich, factually accurate, and grammatically sound localized descriptions, which are critical for geospatial analysis. Although GPT-4o outperforms DescribeEarth in ``Usage'' related questions, DescribeEarth's overall robustness and domain-specific excellence make it a superior choice for Geo-DLC.


\subsection{Ablation Studies}

\subsubsection{Effect of Domain Guidance}

To understand the influence of domain-specific guidance on model performance, we evaluated three distinct settings, as presented in Table~\ref{tab:category}. The feature-level domain guidance setting, which is the default configuration for DescribeEarth, leverages RemoteCLIP features directly integrated into the fusion module, effectively providing a domain-specific prior. The ``No Guidance'' setting omits this prior, relying solely on direct fusion of global and focal features via gated cross-attention. The text-prompt domain guidance setting, in contrast, appends RemoteCLIP predictions as a simple textual hint (``RemoteCLIP suggests it's $<$category$>$'') to the prompt during both training and inference.
The results clearly demonstrate that feature guidance yields the best overall performance, indicating the crucial role of domain-specific categorical priors in enhancing localized captioning. 
Text guidance performs better than ``No Guidance'' on simple instances, suggesting that even a simple textual hint can aid in straightforward recognition tasks. However, in more complex and OOD scenarios, text guidance performs worse than No Domain Guidance. This indicates that directly relying on RemoteCLIP's raw category predictions can introduce misleading information when these predictions are inaccurate for challenging or unfamiliar objects. Our feature guidance strategy, by contrast, integrates RemoteCLIP's visual features more subtly within the fusion module, avoiding direct reliance on potentially erroneous explicit category predictions and thus maintaining robustness across diverse scenarios. 



\begin{table}[t]
\centering
\caption{Comparison of different domain guidance strategies on DE-Benchmark (in \%). Numbers in parentheses denote the performance gap. Positive values in green, negative in red.}
\label{tab:category}
\begin{tabular}{lccc}
\toprule
  & \textbf{Feature Guidance} & No Guidance & Text Guidance \\
\midrule
Simple & \textbf{74.41} & 72.03 (\textcolor{green}{+2.38}) & 73.50 (\textcolor{green}{+0.91}) \\
Complex  & \textbf{70.04} & 68.72 (\textcolor{green}{+1.32}) & 68.43 (\textcolor{green}{+1.61}) \\
OOD & \textbf{66.48} & 65.67 (\textcolor{green}{+0.81}) & 64.90 (\textcolor{green}{+1.58}) \\
\bottomrule
\end{tabular}
\end{table}

\subsubsection{Effect of Feature Fusion Strategy}

Table~\ref{tab:fusion} explores the impact of different feature fusion strategies, with our DFM achieving the highest scores at both complex and OOD levels. In comparison, the pure gated cross attention strategy (denoted as ``GCA'' in Table~\ref{tab:fusion}), which directly fuses global and focal features, performs worse than DFM across all levels (\textit{e.g.}, 72.03\% for simple instances, 2.38\% lower than DFM). The ``Concat'' strategy, which simply concatenates global and focal features, yields 70.31\% for simple instances, appearing competitive with GCA. However, its performance significantly degrades for complex and especially OOD instances, showing a greater decline compared to DFM. This indicates that simple feature concatenation is insufficient to capture the complex relationships and domain-specific insights required for Geo-DLC, particularly when dealing with novel or challenging instances, where deep semantic interaction during fusion is critical. Similarly, employing a Q-Former~\cite{li2023blip} as the fusion module resulted in significantly lower performance than DFM, further emphasizing the efficacy of our DFM design in integrating domain-specific priors to enrich visual representations.




\begin{table}[t]
\centering
\caption{Comparison of different feature fusion strategies. Numbers in parentheses denote the performance gap. Positive values in green, negative in red.}
\label{tab:fusion}
\begin{tabular}{lcccc}
\toprule
  & \textbf{DFM} & GCA & Concat & Q-Former \\
\midrule
Simple & 74.41 & 72.03 (\textcolor{green}{+2.38}) & \textbf{75.20} (\textcolor{red}{-0.79}) & 65.97 (\textcolor{green}{+8.44}) \\
Complex  & \textbf{70.04} & 68.72 (\textcolor{green}{+1.32}) & 69.78 (\textcolor{green}{+0.26}) & 64.82 (\textcolor{green}{+5.22}) \\
OOD & \textbf{66.48} & 65.67 (\textcolor{green}{+0.81}) & 65.07 (\textcolor{green}{+1.41}) & 64.85 (\textcolor{green}{+1.63}) \\
\bottomrule
\end{tabular}
\end{table}

\subsubsection{Effect of Vision Module Training}

We further investigate whether training the vision encoder affects model performance, which is a subtle difference in training between DescribeEarth and DAM. DAM integrate localization information by altering the visual input stream itself, typically by adding a mask as a fourth input channel. This input format necessitates fine-tuning the vision encoder to adapt to the new data distribution. Conversely, DescribeEarth is designed to decouple localization from the visual encoder. We provide location information (\textit{i.e.}, OBB coordinates) textually, directly to the LLM. This approach leaves the visual input pristine and in-distribution for the pre-trained vision encoder. Consequently, we can freeze the vision encoder, preserving its powerful, pre-trained representations without the risk of degradation or over-fitting.
To validate this design choice, we compare two settings: the fully frozen vision encoder and the trainable vision encoder. The results in Table~\ref{tab:visiontrain} affirm our strategy. Freezing the encoder not only enhances generalization to simple (+1.24\%) and OOD (+1.52\%) instances but also maintains the integrity of the pre-trained visual backbone. While a trainable encoder shows a marginal improvement in complex scenarios (0.23\%), the overall benefits of our freezing strategy make it a more parameter-efficient and robust approach for the Geo-DLC task.


\begin{table}[t]
\centering
\caption{Comparison of whether the visual encoder is frozen (in \%). Numbers in parentheses denote the performance gap. Positive values in green, negative in red.}
\label{tab:visiontrain}
\begin{tabular}{llcc}
\toprule
 & \textbf{Frozen} & Trainable \\
\midrule
Simple & \textbf{74.41} & 73.17 (\textcolor{green}{+1.24}) \\
Complex  & 70.04 & \textbf{70.27} (\textcolor{red}{-0.23}) \\
OOD & \textbf{66.48} & 64.96 (\textcolor{green}{+1.52}) \\
\bottomrule
\end{tabular}
\end{table}

\begin{table}[t]
\centering
\caption{Ablation of the focal strategy and global image's input resolution (in \%). Numbers in parentheses denote the performance gap. Positive values in green, negative in red.}
\label{tab:resolution}
\begin{tabular}{lccc}
\toprule
Resolution & \textbf{448$\times$448} & 448$\times$448 & 224$\times$224 \\
\midrule
Focal strategy & \textbf{Scale-adaptive} & Fixed-size & Scale-adaptive \\
\midrule
Simple & 74.19 & 73.59 (\textcolor{green}{+0.60}) & \textbf{74.41} (\textcolor{red}{-0.22}) \\
Complex  & \textbf{70.24} & 69.58 (\textcolor{green}{+0.66}) & 70.04 (\textcolor{green}{+0.20}) \\
OOD & \textbf{68.78} & 66.54 (\textcolor{green}{+2.24}) & 66.48 (\textcolor{green}{+2.30}) \\
\bottomrule
\end{tabular}
\end{table}

\subsubsection{Effect of Scale-Adaptive Focal Strategy}

Our scale-adaptive focal strategy is meticulously designed to balance the encoding of fine-grained target details with essential surrounding context. In this strategy, we employ a hierarchical cropping approach for objects of different sizes to ensure the model perceives both the object itself and its environment. As listed in Table~\ref{tab:resolution}, scale-adaptive focal strategy proves crucial for Geo-DLC, significantly outperforming the fixed-size strategy (directly cropping the object's bounding box). This confirms that dynamically adjusting the focal crop size is critical for accommodating the vast scale variations in remote sensing image.
To further enhance this capability, we investigate the effect of the global image resolution. The results in Table~\ref{tab:resolution} confirm that utilizing a higher-resolution global image ($448 \times 448$) consistently improves performance, showing gains on complex ($+0.20\%$) and OOD instances ($+2.30\%$). This demonstrates that a clearer global context is instrumental for improving the effectiveness of the focal strategy, allowing for more accurate and context-aware descriptions.




\begin{figure}
    \centering
    \includegraphics[width=1\linewidth]{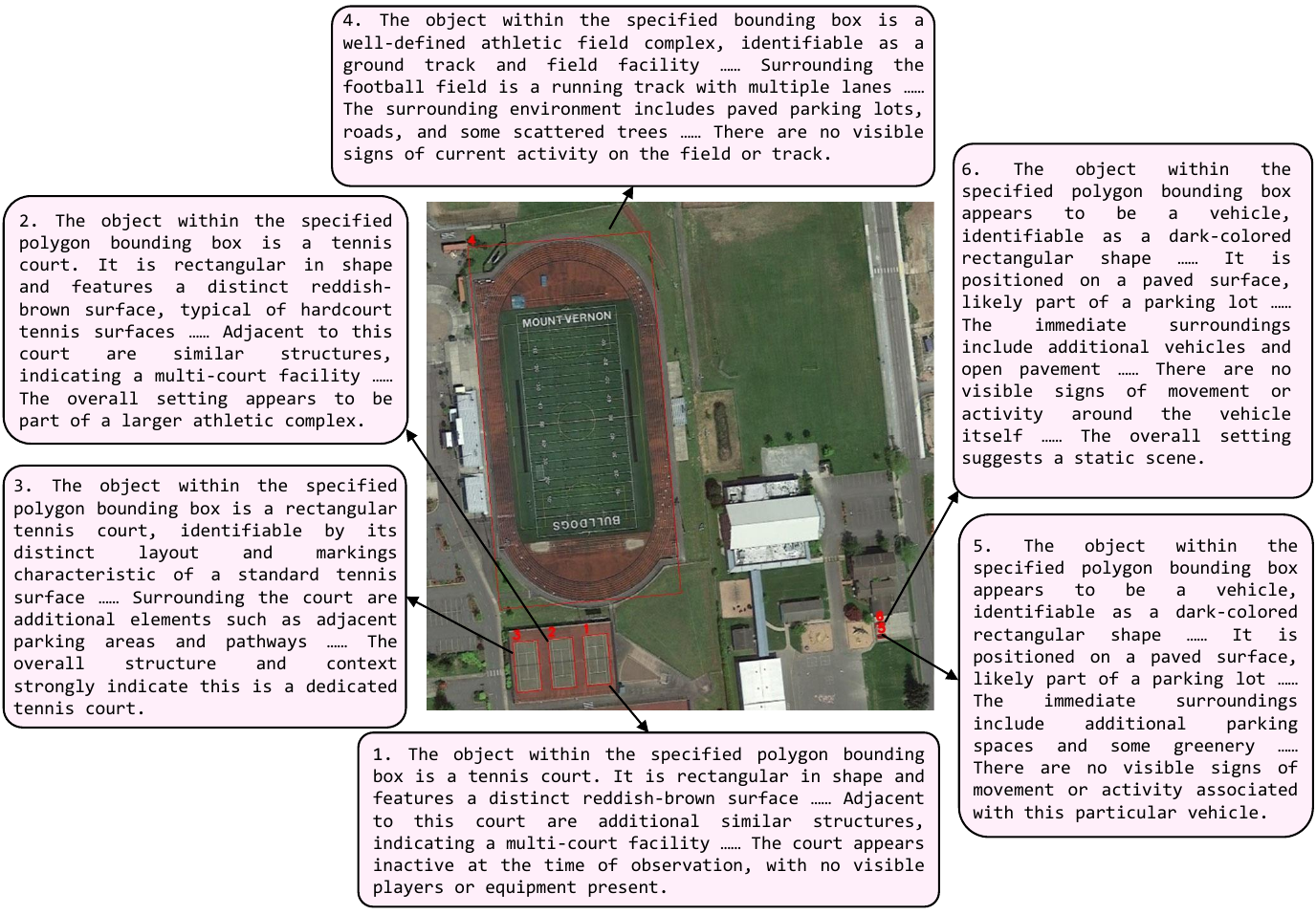}
    \caption{Pipeline for automated image captioning}
    \label{fig:Figure_further_perspectives}
\end{figure}

\subsection{Integration with Detection Models}

Beyond user-interactive queries, a key application of DescribeEarth is its integration into fully automated pipelines for large-scale geospatial analysis. This capability allows for the automatic enrichment of remote sensing image with dense, semantic annotations. As illustrated in Fig.~\ref{fig:Figure_further_perspectives}, the workflow operates in two main stages. First, an off-the-shelf object detection model or region proposal network is applied to a remote sensing image to identify and localize objects of interest, outputting their respective OBBs. Second, each detected OBB, along with the original image, is sequentially fed into our DescribeEarth model, which then generates a detailed, context-aware textual description for the specific object within that box. This integration creates a powerful and scalable solution for automatically annotating vast amounts of remote sensing data. It moves beyond simple class labels, enriching geospatial datasets with semantic, human-readable information that can be indexed and searched, thereby significantly enhancing the capabilities of geographic information systems and enabling advanced analytical queries.

\section{Conclusion}

In this paper, we present a comprehensive framework to advance Geo-DLC, a critical task for the semantic interpretation of remote sensing imagery. To enable meaningful progress, we introduce DE-Dataset, the first large-scale dataset specifically designed for this task, featuring instance-level OBBs paired with rich, fine-grained descriptions. Recognizing the limitations of traditional evaluation methods, we also develop DE-Benchmark, an attribute-based and LLM-assisted protocol that ensures a fair and nuanced assessment of a model's ability to generate factually accurate and contextually relevant captions. These resources are instrumental in the development and validation of our DescribeEarth model, which demonstrates a marked improvement in interpreting complex geospatial objects over state-of-the-art closed-source alternatives. Ultimately, this work establishes a new standard for Geo-DLC and provides a robust foundation to spur the development of more sophisticated vision-language models capable of unlocking deeper insights from Earth observation data for automated analysis, monitoring, and intelligence applications.

\balance

\bibliographystyle{IEEEtran}
\bibliography{ref.bib}

\vfill

\end{document}